\documentclass[
aip,
reprint,
]{revtex4-1}
\usepackage{amsmath, amssymb,amsfonts}
\usepackage{float}
\usepackage{algorithmic}
\usepackage{graphicx}
\usepackage{textcomp}
\usepackage{hyperref}
\usepackage{xcolor}
\usepackage[normalem]{ulem}

\draft 

\begin{document}


\title{Reconstructing shared dynamics with a deep neural network} 



\author{Zsigmond Benk\H{o}}
\email[]{benko.zsigmond@wigner.hu}
\affiliation{Department of Computational Sciences, Wigner RCP, Budapest, Hungary}

\author{Zoltán Somogyvári}
\affiliation{Department of Computational Sciences, Wigner RCP, Budapest, Hungary}


\date{\today}

\begin{abstract}
Determining hidden shared patterns behind dynamic phenomena can be a gamechanger in multiple areas of research.
Here we present the principles and show a method to identify hidden shared dynamics from time series by a two-module, feedforward neural network architecture: the Mapper-Coach network. 
We reconstruct unobserved, continuous latent variable input, the time series generated by a chaotic logistic map, from the observed values of two simultaneously forced chaotic logistic maps.
The network has been trained to predict one of the observed time series based on its own past and conditioned on the other observed time series by error-back propagation.
It was shown, that after this prediction have been learned successfully, the activity of the bottleneck neuron, connecting the mapper and the coach module, correlated strongly with the latent shared input variable.
The method has the potential to reveal hidden components of dynamical systems, where experimental intervention is not possible.
\end{abstract}


\maketitle 

\section{Introduction \& Background}

Theory of nonlinear dynamical systems has plenty to offer for neural architecture design. Here we take a dualistic approach, incorporating the principles of nonlinear time series techniques and the application of simple feedforward learning algorithms to achieve the reconstruction an unobserved common input of two observed chaotic dynamical systems.  

A dynamical system has a state and an update rule. 
It can have a discrete time dynamics or it can be a continuous time flow. 
In the former, the state ($s_{t-1}$) is updated according to the update rule ($f$) thus gives rise to the future states of the system, which we observe through an observation function ($g$, Eq.\,\ref{eq:ds}).
\begin{equation}\label{eq:ds}
    \begin{split}
        s_{t} &= f(s_{t-1})\\
        x_t &= g(s_{t})
    \end{split}
\end{equation}
Where the $s \in \mathbb{R}^p$, $f: \mathbb{R}^p \rightarrow \mathbb{R}^p$ is a smooth (not necessarily invertible) mapping,  $g: \mathbb{R}^p \rightarrow \mathbb{R}^q$ is a smooth observation function which produces the $x_t \in \mathbb{R}^q$ time series ($t \in \{0, 1, 2, ..., T-1, T\}$ ).

\begin{figure}[htb!]
    \centering
    \includegraphics[width=0.5\textwidth]{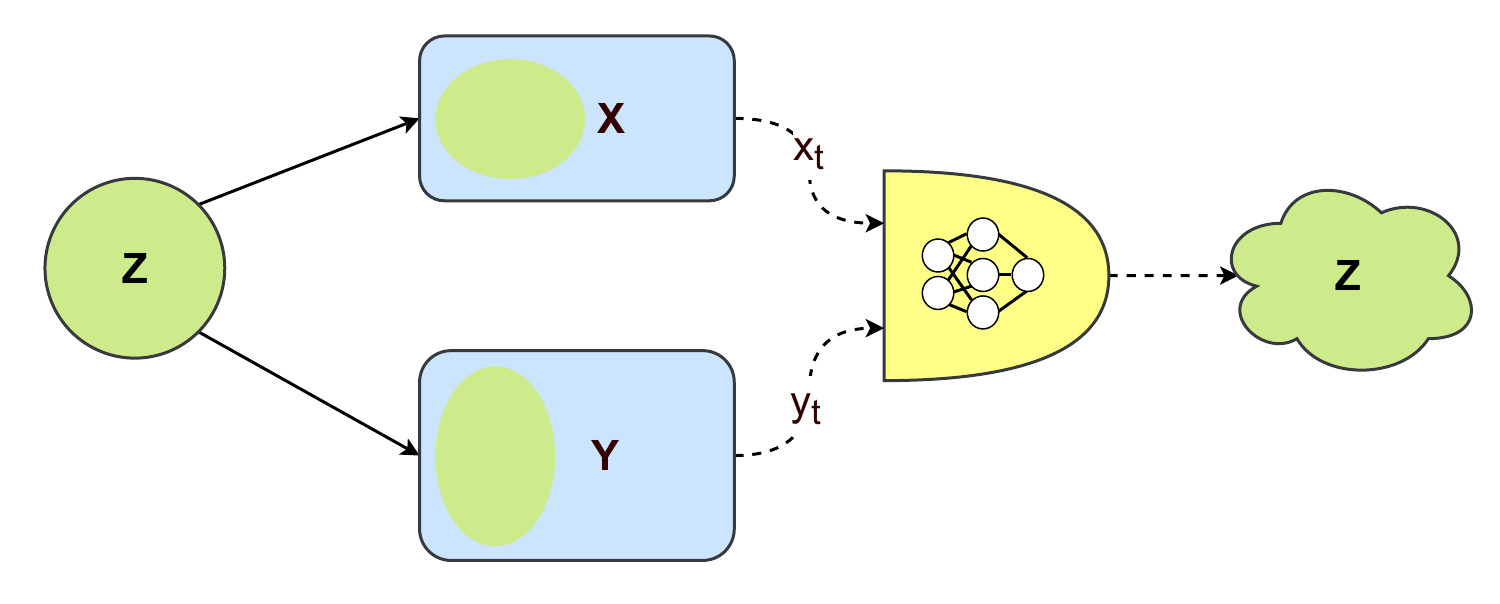}
    \caption{Shared dynamics reconstruction from time series.
    The shared latent driver ($Z$) leaves similar mark (green) on the forced subsystems ($X$ and $Y$).
    This redundancy can be exploited to extract the latent part of the dynamics from time series observations of the forced parts using  artificial neural networks.}
    \label{fig:abstract}
\end{figure}

If we have no acces to the full system state but have an observation function like $g$ producing a time series ($x_t$), we can still reconstruct the state up to a continuous transformation.
Takens theorem\cite{Takens1981} ensures that this topological equivalence has high probability, and shows that time delay embedding -- shown in \eqref{eq:takens} -- is such a reconstruction procedure.
\begin{equation}\label{eq:takens}
    X_t = [x_t, x_{t-1}, \quad \hdots  \quad, x_{t-(2p-1)}]
\end{equation}
Where $X_t$ is the reconstructed state at time $t$.

In the case of directionally coupled systems, one can reconstruct the state of the driver from the state of the forced subsystem \cite{Stark1999,Stark2003,Sugihara2012}. Directed causal link leaves the fingerprint of the driver on the driven system such that the full coupled system's state can be reconstructed or decoded from the observations of the driven system.

When a hidden process drives two subsystems, both subsystems contain a fingerprint of the latent variable that is encoded in their dynamics (Fig.\,\ref{fig:abstract}).
This redundancy can be leveraged to identify the existence hidden common drivers between subsystems \cite{Hirata2010a, Benko2018a}.
Hirata et al.\,applied the technique of recurrence maps to identify the existence of the hidden common cause.
Benko et al.\, measured the intrinsic dimensionality of reconstructed statespaces, thus identified the existence of the hidden common cause by determining the number of the redundant degrees of freedom (the mutual dimension \cite{Sugiyama2013}) between the reconstructed dynamics of the subsystems.

Since the redundant information about the shared hidden dynamics or common cause is present in the reconstructed states of the forced subsystems,
then the question arises whether this information can be extracted from the time series or it is practically unidentifiable.
In this article we show that not only the existence, but values of the shared dynamics can be reconstructed (up to a continuous transformation) by a new specific neural network architecture.

\section{Previous work}

There are only a few methods to reconstruct hidden common input from multivariate time series observations and the use of these is restricted to specific types of hidden common inputs \cite{Sauer2004,Wiskott2002}.

Sauer demonstrated that reconstruction of common input from time series is possible by the Shared Dynamics Algorithm (SDA) if the observed dynamics is simple\cite{Sauer2004, Sauer2010}.
SDA builds on the shared dynamics reconstruction theorem, and makes possible to reconstuct unobserved and shared driver dynamics from time series observations of driven systems[1].

The steps of this family of algorithms are as follows:
\begin{enumerate}
    \item Time delay embedding of the $n$ time series $i=1...n$ with the shared dynamics, $t =1 \dots T$. The embedding dimension is $m$ resulting in reconstructed points $X_i \in \mathbb{R} ^ {T \times m}$.
    \item Pick the first variable ($X_1$) as reference for the equivalence classes. Choose neighborhoods from $X_j | j=2 \dots m$ at times $t$,  and pick $\varepsilon$ radius neighborhoods around.
    Map back the points within the neighborhoods onto $X_1$, and connect the sets which are closer than $\varepsilon$ in $X_1$.
    Cluster the sets and choose representative points from the clusters.
\end{enumerate}

In the end, the algorithm splits the points into equivalence classes, each class stands for a discretized value of the latent driver.

Wiskott and Sejnowski introduced Slow Feature Analysis (SFA) as a Blind Source Separation (BSS) technique \cite{Wiskott2002}.
The objective of BSS is to identify the true signal sources ($S$) from multivariate time series measurements ($X$), which contain the source signals in a mixed way: 
\begin{equation}
X = f(S)    
\end{equation}
where 
$X \in \mathbb{R}^{T\times d}$ is the time series,
$S \in \mathbb{R}^{T\times d}$ is the latent signal,
$S^{(i)}$-s are the source factors,
$f: \mathbb{R} \rightarrow \mathbb{R}$ is the invertible mixing function and the $(f, S)$  pair is a Latent Variable Model.
The BSS problem is underdetermined in general, so additional assumptions on the data and the mixing function is required to find a unique solution.
Such an assumption is that the sources were independent  and non-gaussian like in linear Independent Component Analysis (ICA).
It is also possible to identify independent features in the case of nonlinear mixing functions with Variational Auto-Encoders\cite{Khemakhem2019} and normalizing Flow-Based Methods\cite{Sorrenson2020}.

Slow Feature Analysis (SFA) solves the blind source separation problem for time series\cite{Wiskott2002}.
It assumes that the latent signals are independent and temporally consistent, slowly changing:
\begin{equation}
     \min_{f: X=f(Z)} \sum_i \left< \Delta Z^{(i)}\right>, \quad \left< z \right> = 0, \quad Z^{T}Z=I
\end{equation}
The suitable mapping chosen by applying a filterbank on the original signal then performing PCA  on whitened features to get slow features.
This basic SFA method was implemeted recently using neural networks resulting in the slow flow model\cite{Pineau2020}.

Wiskott showed that one can reconstruct slow hidden common dynamics from time series with SFA\cite{Wiskott2003}. 

However, no method existed yet, which can reconstruct a continuous hidden common input with fast dynamics.    
Here we introduce such a method in order to reconstruct fast and continuous latent shared dynamics from time series.

\begin{figure}[tb!]
\centerline{\includegraphics[scale=0.55]{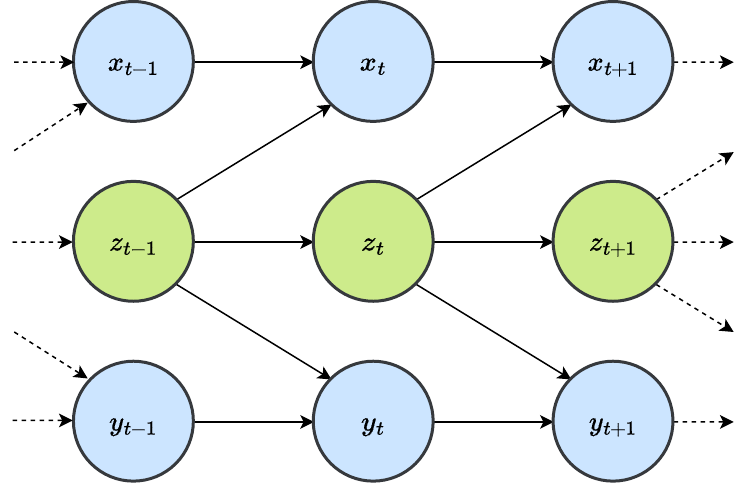}}
\caption{Directed graph representation of the causal connections in the coupled logistic map system.
The $z$ variable (red) is a hidden common cause of the two observed variables, $x$ and $y$ (blue). }
\label{fig:ts}
\end{figure}

\section{Example: coupled logistic map system}

We demonstrate the nonlinear state space reconstruction and the inference procedure for the latent process on  coupled logistic map systems.
The logistic map is a discrete time dynamical system exhibiting chaotic behaviour in a properly selected range of the parameter $r$\cite{May1976}.

Let's take three logistic maps coupled in a fork structure such that one drove the other two (Fig.\,\ref{fig:ts}, Eq.\,\ref{eq:logmap}).
\begin{equation}\label{eq:logmap}
    \begin{split}
            x_t &= r x_{t-1} (1 - x_{t-1} - \beta_{x z} z_{t-1}) \\
            y_t &= r  y_{t-1} (1 - y_{t-1} - \beta_{y z} z_{t-1}) \\
             z_t &= r  z_{t-1} (1 - z_{t-1})
    \end{split}
\end{equation}
Where $r$ is a positive parameter and $\beta_{ij}$ are coupling constants.

Let's focus on the $y$-$z$ subsystem, where $z$  unidirectionally drives $y$.
The state of the subsystem $S_t$ is the $(z_t, y_t)^T$ vector and the dynamics of the system can be written in the corresponding vector form:
\begin{equation}\label{eq:loguni_dynamics}
    \begin{split}
        S_t &= F(S_{t-1}) = 
        \left(
        \begin{array}{c}
            r  z_{t-1} (1 - z_{t-1})\\
            r y_{t-1} (1 - y_{t-1} - \beta_{y z} z_{t-1})\\
        \end{array}
        \right)\\
        y_t &= g(S_t)
    \end{split}
\end{equation}
where $y$ and $z$ are the state variables, $r$ is the parameter of the logistic map and $\beta_{ij}$ are the coupling parameters.
Furthermore, $g: \mathbb{R}^2 \to \mathbb{R}$ is a linear observation function such that $g(S_t) = y_t$. 
So $y_t$ values are observed from the system state, but the $z_t$ values remain hidden.

\begin{figure}[tb!]
\centerline{\includegraphics[scale=0.55]{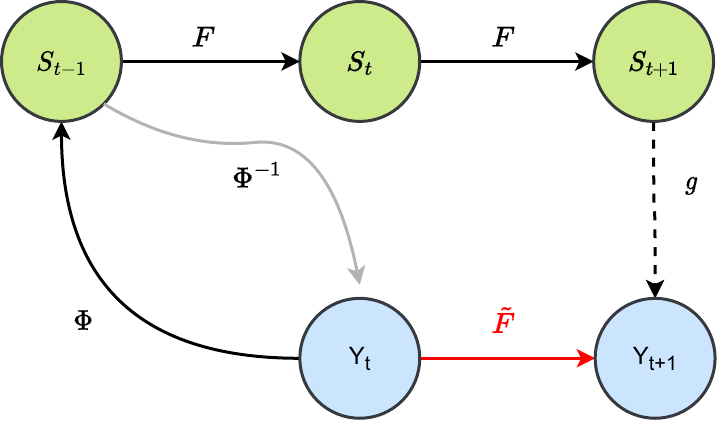}}
\caption{Dynamics on the reconstructed state space.
The next value $y_{t+1}$ and state can be written as a function of current reconstructed state following the black arrows from $Y_t$ to $Y_{t+1}$.
The composition of mappings along the route realize the $\Tilde{F}$ dynamics on the reconstructed space.
The $\Phi$ embedding function is invertible, however the dynamics $F$ is not invertible, neither the observation function $g$.
}
\label{fig:dynamics}
\end{figure}

\begin{figure}[tb!]
    \centering
    \includegraphics[scale=0.5]{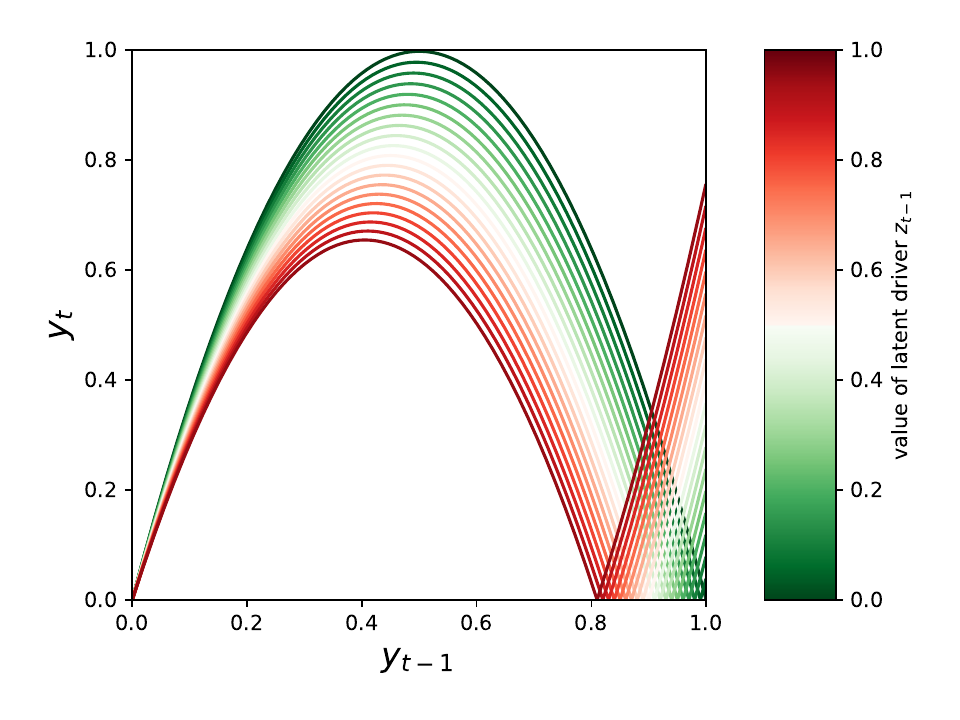}
        \caption{The return map of the $y$ variable.
        The distinct fibers corresponds to different values of the latent variable $z$.
        Due to the coupling and the reflecting boundary conditions, there is a small self-intersecting area in this 2D embedding, where the value of the latent is ambiguous.}
    \label{fig:return_map}
\end{figure}

In the followings, we show $z_t$ can be reconstructed from one of the observed time series.
We know from the theory \cite{Stark1999} that we can reconstruct the state of the subsystem based on the observed variable $y$ time delay embedding:
\begin{equation}
    Y_t = [y_t, y_{t-1}]^T
\end{equation}
where $y_t$ is the scalar value of $y$ at time $t$, and $Y_t \in \mathbb{R}^2$ is the reconstructed state of the system almost everywhere .
Here we neglect the ambiguity caused by the reflective boundary conditions needed to constrain the coupled dynamics in the $[0, 1] \times [0, 1]$ square.
This ambiguity can be ruled out by higher dimensional embeddings for example in $2d+1=5$ dimensions the space would be an embedding of the $2$-dimensional statespace \cite{Takens1981}.
To see that $Y_t$ truly described the state of the system almost everywhere, we show the mapping from $Y_t$  to $S_t$ and build the dynamics in the reconstructed state space (Fig.\,\ref{fig:dynamics}).

First, we reorder the equation that realizes the $y_t = r y_{t-1} (1 - y_{t-1} - \beta_{y z} z_{t-1})$ mapping (Eq.\,\eqref{eq:loguni_dynamics}) to get the expression for $z_{t-1}$ and $S_{t-1}$: 
\begin{equation}\label{eq:phi}
 S_{t-1} = \Phi(Y_t) = 
  \begin{pmatrix}
       \phi_1(Y_t)\\
       \phi_2(Y_t)
 \end{pmatrix}
 =
 \begin{pmatrix}
       z_{t-1}\\
       y_{t-1}
 \end{pmatrix}
 =
\begin{pmatrix}
            \frac{1 - y_{t-1} - \frac{y_t}{r  y_{t-1}}}{\beta_{yz}}\\
      y_{t-1}
\end{pmatrix}
\end{equation}
thus we have a mapping from $Y$ to $S$.

Also, we can write the the inverse of $\Phi$ as:
\begin{equation}
    \Phi^{-1}(S_{t-1}) = Y_t = 
    \begin{pmatrix}
      r y_{t-1} (1 - y_{t-1} - \beta_{yz} z_{t-1})\\
      y_{t-1}
    \end{pmatrix}
\end{equation}
By this, we demonstrated that there exists a one-to-one mapping between $Y_t$ and $(z_{t-1}, y_{t-1})$, thus $Y$ is a reconstructed state of the $y-z$ subsystem.
We can define the dynamics $\Tilde{F}$ on this reconstructed space, to predict future states (Fig.\,\ref{fig:dynamics}).
\begin{equation}
    Y_{t+1} = \Tilde{F}(Y_t) = 
    \begin{pmatrix}
          g \circ F \circ F \circ \Phi(Y) \\
          y_t
    \end{pmatrix}
\end{equation}

Let's zoom out from the $y-z$ subsystem and see the full $x-y-z$ system, where $z$ is unobserved.
To predict the value of $x_t$, one needs the previous value of $x_{t-1}$ and $z_{t-1}$ according to the update rule in Eq.\,\eqref{eq:logmap}.
Since $Y$ contains all required information about $z$, we can use $Y_t$ in place of $z_{t-1}$, and predict $x_t$ given $x_{t-1}$, $Y_t$ if the required mappings - $\Phi$, $F$ and $g$ -  are also given.
When these functions are unkown, one has to approximate them from data.
For the function approximation, it is practical to use feedforward neural networks.

\begin{figure}[tb!]
\centerline{\includegraphics[scale=0.3]{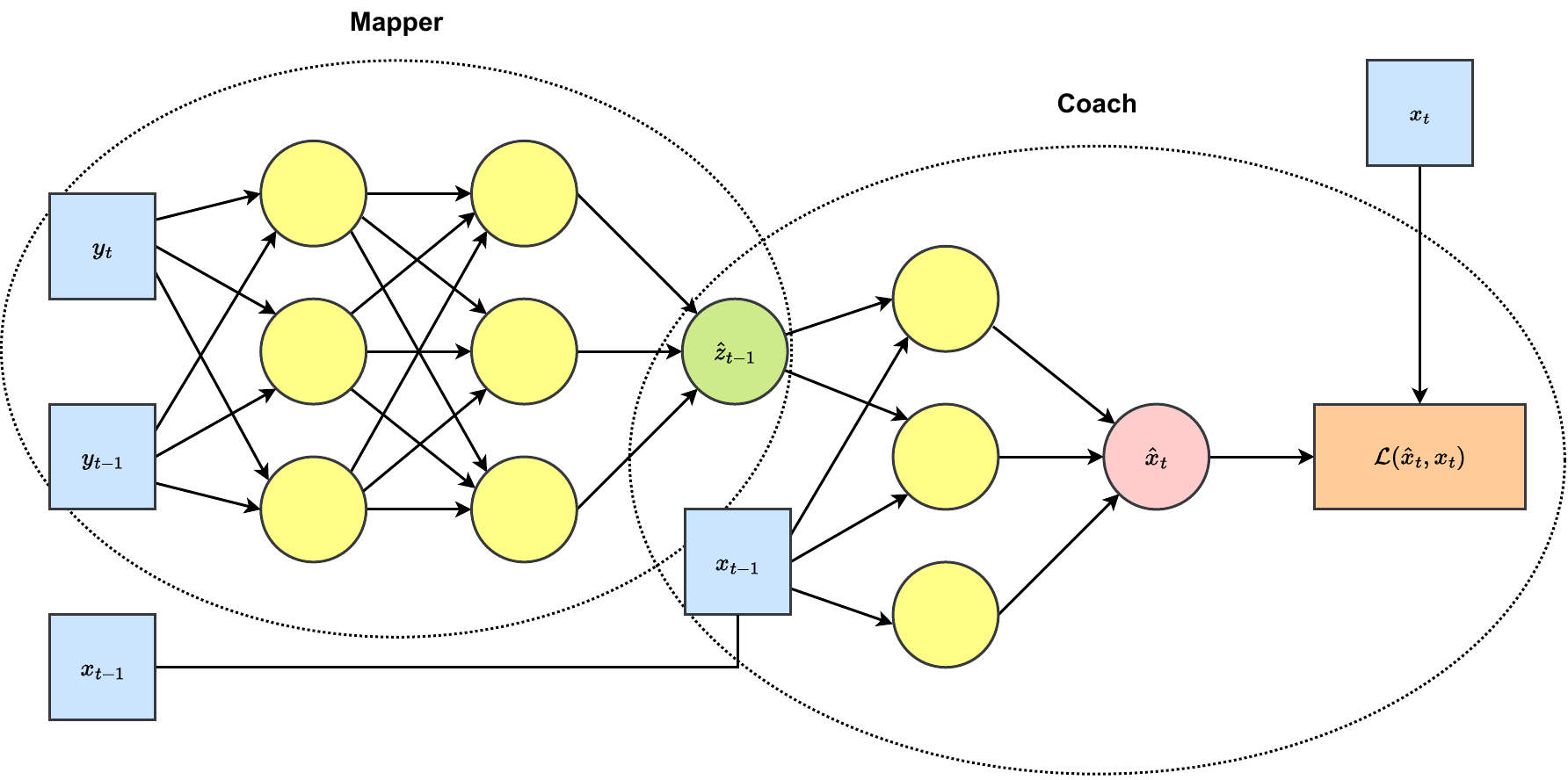}}
\caption{The architecture of the mapper-coach network.
The feedforward neural network predicts present values of $x$.
One cannot predict accurately $x(t)$ values based on solely $x(t-1)$, because the lack of information about $z(t-1)$.
This missing information is sipped out from $Y(t)$ through the red bottleneck
At the training phase the full mapper-coach network can be trained with backpropagation algorithm and -- after successful training -- the activity of the red hidden unit has a one-to-one correspondence with values of the latent $z$ variable.
At this reconstruction stage, the coach can be detached from the mapper, which latter estimates the mapping ($\phi$) from the reconstructed state space of the effect ($Y$) to the state space of the hidden variable ($z$).}
\label{fig:complex}
\end{figure}

\section{Results}

\subsection{The proposed architecture: the Mapper-Coach network}

We reconstruct the hidden common input from a dataset generated by the logistic map system in Eq.\,\eqref{eq:logmap}.
To extract the $z$ hidden driver, we set up a prediction task and a neural network architecture for function approximation (Fig.\,\ref{fig:complex}).
We predict the value of $x_t$ based on  $[x_{t-1}, Y_{t}]$ as input. 
We split this task into two parts: first we estimate $z_{t-1}$ from $Y_t$ and second we approximate the dynamics from the estimated state. 

To estimate the $z_{t-1}$ hidden input we had to approximate the $\phi_1$ mapping (Eq.\eqref{eq:phi}), we implemented this function by a feedforward neural network: the Mapper module (Fig.\,\ref{fig:complex}).
The input space is the reconstructed state space of $Y$  and the output is the value of $z$. 
But how to find the exact mapping from $Y$ to $z$, when $z$ is unobserved?

We use values from the other observed time series $x_t$ to guide the search for a good mapping.
We attached a coach module (coach) to the output of the mapper and also channeled $x_{t-1}$ as additional direct input (Fig.\,\ref{fig:complex}).
The output of the coach is a prediction on the current value of $x$, therefore the subnetwork intends to implement the dynamics of $x$ from Eq.\,\eqref{eq:logmap}.

The fusion of the two parts is the Mapper-Coach network, which reconstructs the shared dynamics by the activity of hidden units at the meeting point of the two modules.

\begin{figure}[tb!]
\centerline{\includegraphics[scale=0.55]{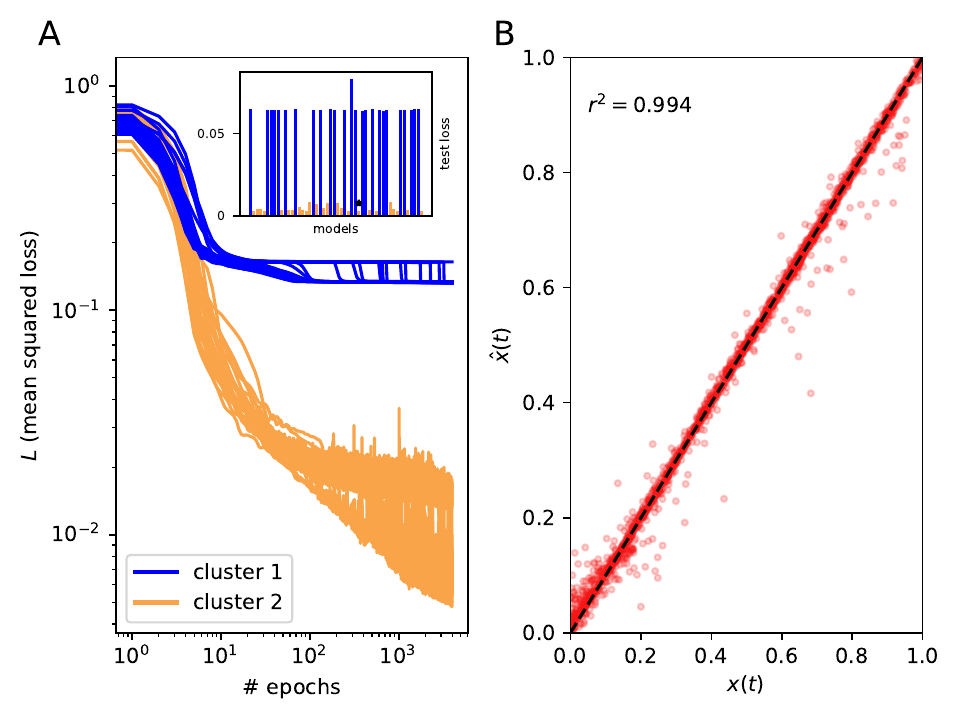}}
\caption{Learning and prediction performance of the network.
{\bf A} Learning curves. The learning-process of fifty mapper-coach networks is shown on the left. 25 instances could not learn and 25 instances converged to smaller mean squared loss values.
{\bf B} Prediction performance of the best model on the test set. The correlogram shows the actual values against the predicted ones.
The point-cloud nicely lies around the diagonal, the coefficient of determination is very high ($r^2=0.99$).
}
\label{fig:forecast}
\end{figure}

\subsection{Performance on the logistic map system}

We trained the mapper-coach network with the backpropagation algorithm to predict $x_t$ at the output of the coach module, and found that the output of the mapper network reconstructed the hidden variable.

{\bf{Prediction performance: }}
The network produced very accurate predictions on $x_t$ after the learning procedure ($r^2=0.99$, Fig. \ref{fig:forecast}).
We split the data into training, test  and validation sets ($n=20k$, $80\%$, $10\%$, $10\%$ respectively), trained the network and evaluated the results on the latter sets.
We reinitialized the training $50$ times with random weights, and selected the best-performing model for evaluation.
According to the learning characteristics, we identified two clusters of the models (\ref{fig:forecast}\,A).
Half of the models were stuck in local minima (cluster 1, 25/50) and in half of the cases the learning was successful (cluster 2, 25/50). 

\begin{figure}[tb!]
\centerline{\includegraphics[scale=0.55]{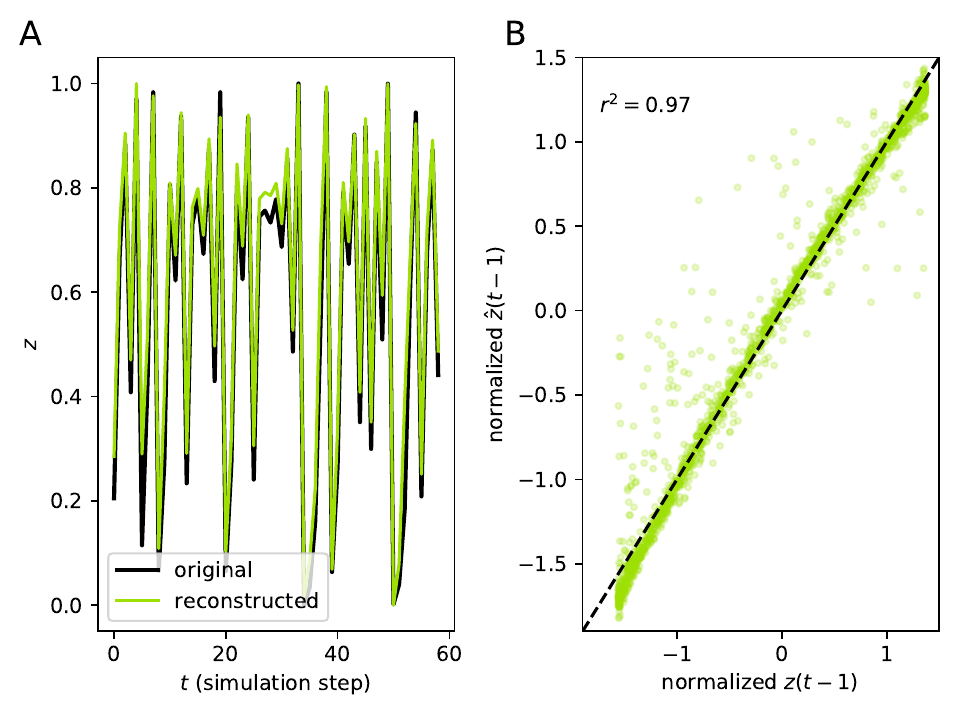}}
\caption{The output and performance of the mapper in reconstruction-mode on the test set.
{\bf A} Traces of original unobserved time series (black) and reconstructed values (red). After reversing and rescaling the amplitudes, the values are almost identical. 
{\bf B} Reconstruction performance on the test set. The subplot shows the original values against the reconstructed ones. The points lie around the diagonal, thus the reconstruction has high quality ($r^2=0.97$). 
}
\label{fig:reconstruct}
\end{figure}

{\textbf{Reconstruction performance: }}
We found that the mapper part of the network reconstructed the past values of the hidden common input($z_{t-1}$) quiet accurately: the squared correlation between the mapper output ($\hat{z}_{t-1}$) and the actual hidden common input was $r^2=0.97$, (Fig\,\ref{fig:reconstruct}).
We detached the coach and investigated the output of the mapper by providing $Y(t)$ test inputs.
The output reflected the values of $z(t-1)$, so the mapper network approximately implemented the $\phi_1$ mapping in Eq.\,\eqref{eq:phi}.

\begin{figure}[b!]
\centerline{\includegraphics[scale=0.55]{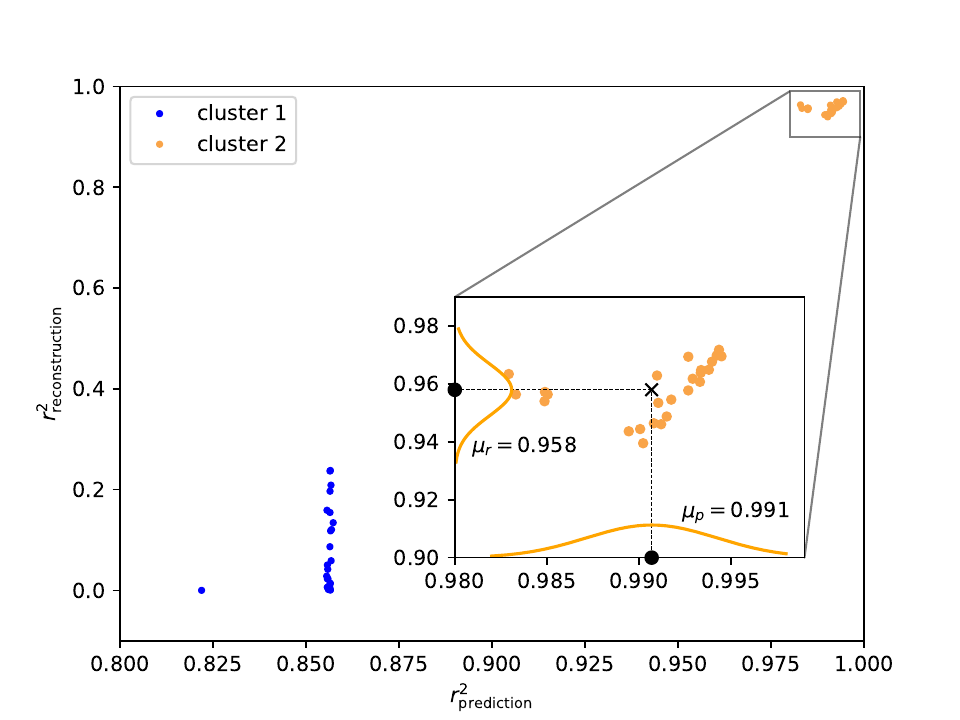}}
\caption{Prediction performance against the quality of reconstruction.
The figure shows the coefficient of variation metrics for different model-instances from the learning procedure.
Reconstruction quality of $z$ is correlated with prediction performance on $x$.
Only the best predictors achieve good reconstruction of the latent variable.
}
\label{fig:mapmaco}
\end{figure}

We showed that the prediction- and the reconstruction performance were related: the better the model predicted $x$ the better it reconstructed $z$ (Fig.\,\ref{fig:mapmaco}).
The trained network instances achieved different prediction performance on the test set, $25$ of the $50$ networks had high quality output.
We plotted the prediction performance against the reconstruction performance for each model.
We saw that better quality predictions implied better quality reconstructions (Fig.\,\ref{fig:mapmaco}).

\section{Discussion}

We demonstrated that reconstruction of a latent variable with fast and continuous state-space dynamics is possible with the backpropagation algorithm, but there are several challenges remain to be assessed.

SDA and SFA can reconstruct common drivers, but the usecases are restricted.
In principle the SDA could be applied on elaborate systems, when the driver and observed dynamics is not discrete or one-dimensional and even for chaotic driver, but in practice it was only demonstrated to work on very simple discrete and one-dimensional examples.
SFA can reconstruct the common latent input, but only if the its dynamics is slow.
In contrast, the mapper-coach approach could reconstruct fast continuous variables.

We reconstructed the latent common input from example time series, leveraging the theoretical properties of the underlying coupled chaotic dynamical systems.
We implemented the method by the Mapper-Coach network, where the mapper part realized the mapping $z_{t-1} = \phi_1(Y)$ and the coach implemented the one-step prediction-map $x_t = f\left( x_{t-1}, z_{t-1} \right)$.

The architecture of the network is asymmetric: the role of the $Y$ and $x$ input is not equivalent.
The reconstructed space of $Y$ contains all information about the hidden variable $z$, but the $x_{t-1}$ value lacks this information.
The vital point is that the coach sips out the required information about $z$ from the $Y$ input through the Mapper, and this information is only fully available in the $Y$, not in the $x$ input pathway.

To ensure that the mapper outputs information about the hidden variable, the dimensionality of the coach-input has to be minimal.
In more complex systems, it could be more difficult to obtain this minimality.
One possibility is that the number of output units from the mapper should equal to dimensionality of the common hidden dynamics.
Also, the number of direct input units (from $x$ in the example) has to equal to the remaining degrees of freedom in $X$ which is independent from $Y$.
Therefore, in this case, to determine the proper number of units for mapper output and $x$-input, one has to investigate the intrinsic dimensionality and the mutual dimension between the embedded $X$, $Y$ variables\cite{Benko2020a,Romano2016, Benko2018a}.
An other possibility would be to use sparse activity-regularization and cross-validation to restrict the information-capacity of the representational layer resulting in a bottleneck.

We demonstrated, that the proposed architecture works for a discrete time, deterministic (noise-free) dynamical system, but whether this approach is applicable to continuous flows and how robust it is to noise is yet to be investigated.
Theoretically there are no obstacles for the former and -- since state-space reconstruction can be robust up to moderate noise levels\cite{Casdagli1991} -- the latter can be handled in the case of low noise, or after noise-filtering.

This architecture gives a simple example on how to extract the information about the common latent variable, but other symmetric, autoencoder-based architectures are also applicable and can be the subject of further investigations.

Also, a possible research-path is the slow-flow realization of shared dynamics reconstruction which is yet to be implemented and tested.

We took a dynamical systems aspect through the paper, but there are connections to autoencoders, representation learning and hidden Markov models\cite{Kingma2014,Marino2018}.
For example the reconstruction of the common dynamics by the proposed architecture can be seen as amortized inference in a  conditional latent variable model.
The coach part, which performs the prediction is the latent variable model conditioned on the $x_{t-1}$ value and the mapper part realizes the fast inference.
Besides the fact that common latent inputs can be reconstructed,  it is relevant to inspect that how the learned latent representations can be interpreted and used for prediction, compression, classification, anomaly detection, etc. \cite{Jakovac2020, Jakovac2021, Jakovac2022,Kurbucz2022}.

\section{Methods}

\subsection{Data generation and handling}
We generated $N=20000$ data points according to Eq.\,\eqref{eq:logmap} ($r=3.99$, $\beta=0.2$) with reflective boundary condition on the $[0, 1]$ interval.
We embedded the $y$ variable, thus produced $Y(t)= [y(t), y(t-1)]$, and aligned the time-scale of $x(t)$ and $z(t)$ time series according to the time-shift produced by the embedding procedure.
We splitted the data into training(80\%), test (10\%) and validation (10\%) sets.

\subsection{Model implementation}

We implemented the mapper-coach network (Fig.\,\ref{fig:complex}) in python\cite{python} with the pyTorch\cite{paszke2017automatic} framework.
For the sake of simplicity we chose the number of hidden units to be the same for each hidden layer ($n=20$), we used rectified linear activation functions.

\subsection{Training and evaluation}
We randomly initialized the weights and started training with the batch size of $2000$ and for $4000$ epochs.
We used the ADAM optimizer and mean squared loss function with default parameters except the learning rate ($10^{-2}$).
We initialized $50$ instances of the network and selected the best performing model on the test set for further evaluation.

We used the coefficient of determination ($r^2$) to evaluate model test and validation performance.
This metric is simply is the square of the Pearson correlation coefficient ($r$):
\begin{equation}
    r_{xy} = \sqrt{\frac{\sum_{i=1} ^{N} (x_i - \mu_x) (y_i - \mu_y)}{\sigma_x \sigma_y}} 
\end{equation}
where $x_i$ and $y_i$ are the sample points, $\mu_j$ are the sample means and the $\sigma_j$-s are the sample standard deviations.
This coefficient describes the linear dependence between the two variables.

We computed $r^2$ between $x(t)$ and the predicted values $\hat{x}(t)$ on the test set (Fig.\,\ref{fig:forecast}).
We have chosen the model with the best performance to see the hidden activations on the validation set.

We also calculated the $r^2$ between the "true" hidden variable $z(t-1)$ and the the output of the mapper network ($\hat{z}(t-1)$) on the validation set.
We displayed these results on a time series plot an on a correlogram (Fig.\,\ref{fig:reconstruct}).

We also computed the reconstructed $z(t-1)$ values and the $r^2$ values for the not optimal models, and plotted the prediction performance against the reconstruction performance of the models (Fig.\,\ref{fig:mapmaco}). 

\section*{Data and Code availability}

The data set and all the code can be found at \href{https://github.com/phrenico/maco_commondriver}{https://github.com/phrenico/maco\_commondriver}.

\section*{Acknowledgments}

The authors thank András Telcs and Marcell Stippinger for the valuable discussions on this topic.
 
The research were supported by the Hungarian National Research, Development, and Innovation Office NKFIH, under grant number K135837 and by the Hungarian Research Network, HUN-REN under grant numbers SA-114/2021 and TECH 2024-20.

\bibliography{ biblio.bib}

\end{document}